\pdfoutput=1

\documentclass[11pt]{article}

\usepackage{ACL2023}
\usepackage{graphicx}
\usepackage{times}
\usepackage{latexsym}

\usepackage[T1]{fontenc}

\usepackage[utf8]{inputenc}

\usepackage{microtype}

\usepackage{inconsolata}

\usepackage{amsmath}
\usepackage{multirow}
\usepackage{booktabs}
\usepackage{colortbl}
\usepackage{amssymb}
\usepackage{bbding}
\usepackage{pifont}
\usepackage{wasysym}
\usepackage{utfsym}
\usepackage{fontawesome}
\usepackage{siunitx}
\usepackage{soul}
\usepackage{pifont}
\definecolor{blue}{HTML}{99CCFF}

\newcommand{\bluehighlight}[1]{\sethlcolor{blue}\hl{#1}}
\definecolor{red}{HTML}{F1C0BE}

\newcommand{\redhighlight}[1]{\sethlcolor{red}\hl{#1}}

\definecolor{green-black}{HTML}{16AB00}
\newcommand{\gtext}[1]{\textcolor{green-black}{#1}}
\definecolor{red1}{HTML}{FF0000}
\newcommand{\rtext}[1]{\textcolor{red1}{#1}}
\title{Improve Student's Reasoning Generalizability\\through Cascading Decomposed CoTs Distillation}

\author{
  Chengwei Dai$^{1,2}$, Kun Li$^{1}$\Thanks{\ Kun Li is the corresponding author.}\ , Wei Zhou$^{1}$, Songlin Hu$^{1}$ \\
  $^{1}$Institute of Information Engineering, Chinese Academy of Sciences\\
  $^{2}$School of Cyber Security, University of Chinese Academy of Sciences\\
  \texttt{\{daichengwei, likun2, zhouwei, husonglin\}@iie.ac.cn}
}
\begin{document}
\maketitle
\begin{abstract}
Large language models (LLMs) exhibit enhanced reasoning at larger scales, driving efforts to distill these capabilities into smaller models via teacher-student learning.
Previous works simply fine-tune student models on teachers' generated Chain-of-Thoughts (CoTs) data. Although these methods enhance in-domain (IND) reasoning performance, they struggle to generalize to out-of-domain (OOD) tasks.
We believe that the widespread spurious correlations between questions and answers may lead the model to preset a specific answer which restricts the diversity and generalizability of its reasoning process.
In this paper, we propose \textbf{Cas}cading Decomposed \textbf{Co}Ts \textbf{D}istillation (CasCoD) to address these issues by decomposing the traditional single-step learning process into two cascaded learning steps. 
Specifically, by restructuring the training objectives—removing the answer from outputs and concatenating the question with the rationale as input—CasCoD's two-step learning process ensures that students focus on learning rationales without interference from the preset answers, thus improving reasoning generalizability. 
Extensive experiments demonstrate the effectiveness of CasCoD on both IND and OOD benchmark reasoning datasets\footnote{Code can be found at \url{https://github.com/C-W-D/CasCoD}}.
\end{abstract}

\section{Introduction}
\label{intro}
Recent developments in LLMs have brought remarkable improvements in multi-hop reasoning tasks via CoT prompting \cite{chain-of-thought-prompting}. However, these great reasoning capabilities are often associated with more parameters \cite{emergent-ability}, which is not practical to emergent in smaller language models (SLMs).
\begin{figure}[htb]
    \centering
    \includegraphics[width=\linewidth]{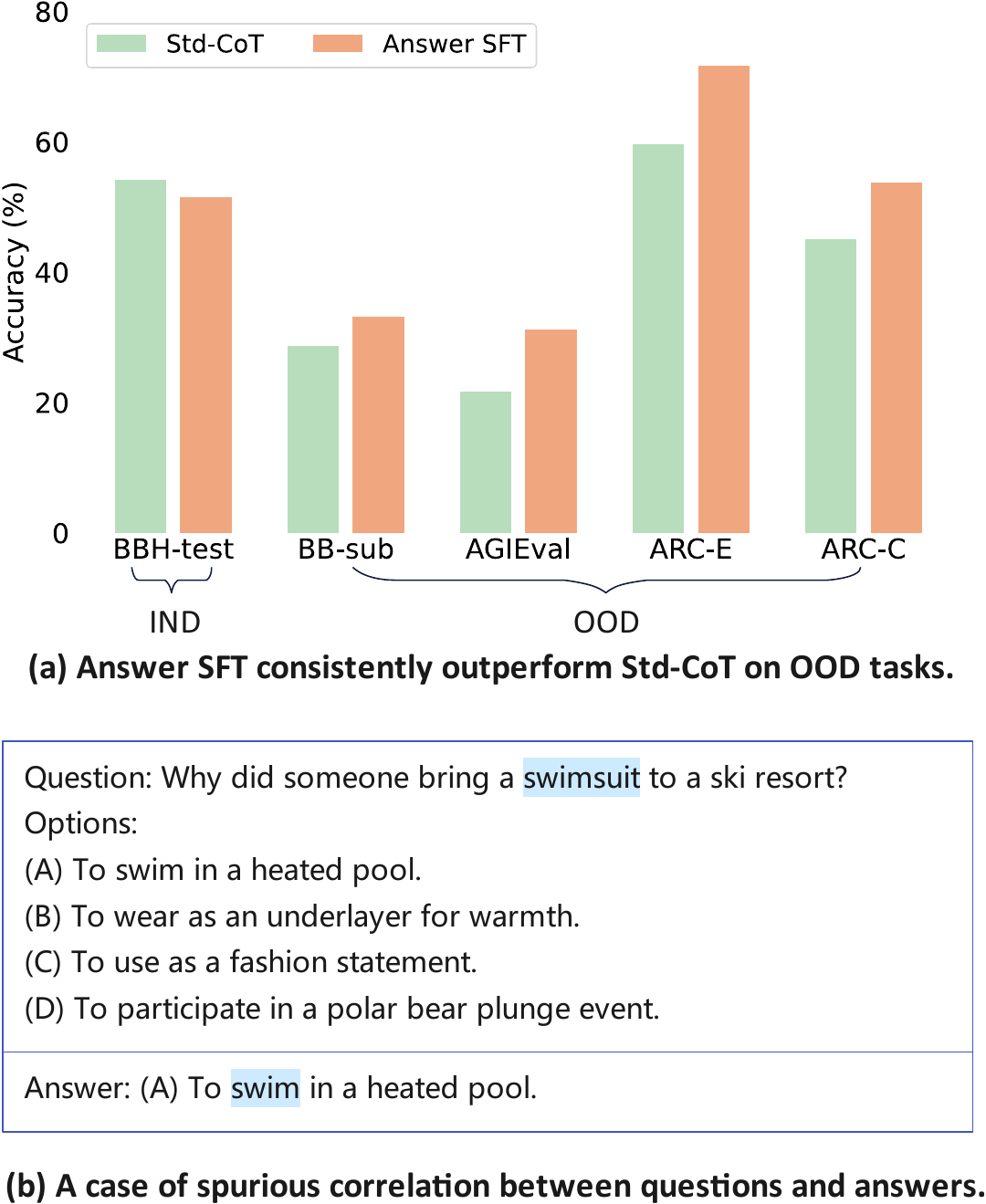}
    \caption{(a) Empirical results of standard CoT distillation (Std-CoT) and directly fine-tuning on answer labels without CoTs (Answer SFT) on one in-domain (BBH-test) and the other four out-of-domain benchmark reasoning datasets. (b) In the given example, the semantic similarity between "swimsuit" in the question and "swim" in the answer demonstrates a high level of match, which could allow the model to predict the answer using simple keyword matching or certain rules.}
    \label{fig:illustration}
\end{figure}

Existing works \cite{acl-teach-slm-reason, acl-lm-are-reasoning-teachers, specializing-slm} try to make the reasoning capabilities isolated and distilled to student SLMs by simply fine-tuning on teacher LLMs generated CoTs data, known as standard CoTs distillation (Std-CoT).
Although the method effectively leverages the LLMs' CoTs to boost the reasoning performance of student models on seen tasks, it does not ensure effective reasoning in OOD settings, leading to weak generalization on unseen tasks.
Our pioneer study demonstrates that, as shown in Figure \ref{fig:illustration} (a), when using the same IND training dataset, student models developed via the method Std-CoT perform better on IND tasks but significantly worse on OOD tasks compared to models fine-tuned directly with question-answer pairs.
The surprising findings indicate that CoTs produced by SLMs do not effectively transfer to new domains and these SLMs seem to be more adept at learning to predict answers directly from questions.

We attribute these issues to the spurious correlations between questions and answers that are commonly found in implicit reasoning tasks \cite{spurious-between-qa-Gururangan, spurious-between-qa-Zellers, spurious-between-qa-Blodgett}, as illustrated in Figure \ref{fig:illustration} (b). The Std-CoT approach requires models to learn both the rationale and the answer in a single step, where the learned spurious correlations in training stage can adversely affect the quality of rationale generation during inference. That is to say, upon reading a question, student models may fastly, unconsciously, and automatically formulate a "preset answer" \cite{thinking-fast-and-slow-in-llm}, which in turn may lead them to implicitly reduce the token generation space when producing CoT. This results in diminished diversity and generalizability of their rationales (intermediate reasoning process).

In this paper, we propose \textbf{Cas}cading decomposed \textbf{Co}Ts \textbf{D}istillation (CasCoD), a straightforward yet effective method to address these issues. 
Specifically, we decompose the traditional single-step learning process of Std-CoT into two cascaded learning steps: a rationale learning step and an answer learning step.
In the rationale learning step, the training objective, with the answer removed, is defined as \footnote{$q$, $r$, and $a$ represent the question, rationale, and answer, respectively.}: $q \rightarrow r$. In the answer learning step, we concatenate the question with the target output from the rationale learning step and use this combined input for the answer learning step, setting the training objective as $q, r \rightarrow a$.
This cascading two-step learning configuration mitigates the capture of spurious correlations between questions and answers during the training phase, ensuring that students focus on learning rationales without interference from the preset answers. Furthermore, the inference phase execution pipeline is aligned with the training phase; the model first generates a rationale when given a question, and then, based on the question-rationale pair, predicts the final answer, further alleviating potential reasoning biases caused by spurious correlations.

Extensive experiments demonstrate that CasCoD outperforms the baselines on both IND and OOD benchmark reasoning datasets. Besides, we validate the generalizability of CasCoD across different model sizes and training data sizes. Further analyses confirm the significant impact of cascading two-step learning and the robustness of CasCoD on OOD tasks. The experiments on reasoning faithfulness and case studies indicate that models distilled by CasCoD can reason more consistently and demonstrate better generalization than baselines, effectively addressing interference from question-answer spurious correlations. Our contributions can be summarized as follows:
\begin{itemize}
    \item We find that standard CoT distillation methods exhibit limited generalizability on OOD tasks, almost performing worse than methods fine-tuned directly with question-answer pairs.
    \item We decompose the traditional single-step learning process into two cascading learning steps to alleviate the impact of spurious correlations between questions and answers.
    \item Extensive experiments confirm the effectiveness of our method across both IND and OOD datasets, showing that CasCoD can generate more generalizable CoTs.
\end{itemize}

\section{Related Works}
\label{sec_related}
\paragraph{CoT Capability of Language Models}
LLMs have demonstrated a wide array of capabilities in numerous natural language processing tasks, underscored by various studies \cite{palm, emergent-ability}. One notable manifestation of this is the CoT prompting technique \cite{chain-of-thought-prompting}, which facilitates models in articulating a series of deductive reasoning steps. This method has substantially enhanced LLMs' problem-solving abilities, as evidenced in several works \cite{let-s-think-step-by-step, self-consistency, language-models-can-self-improve}. Despite these advancements, the effectiveness of CoT prompting notably diminishes in smaller models \cite{emergent-ability}. Research by \citet{flan-t5} indicates that with targeted training on CoT data via instruction tuning, SLMs can unlock CoT capabilities. In our study, we show that SLMs' CoT performance can be further enhanced by decomposing the standard CoT distillation process into two cascaded learning steps. 
\begin{figure*}[t]
    \centering
    \includegraphics[width=\linewidth]{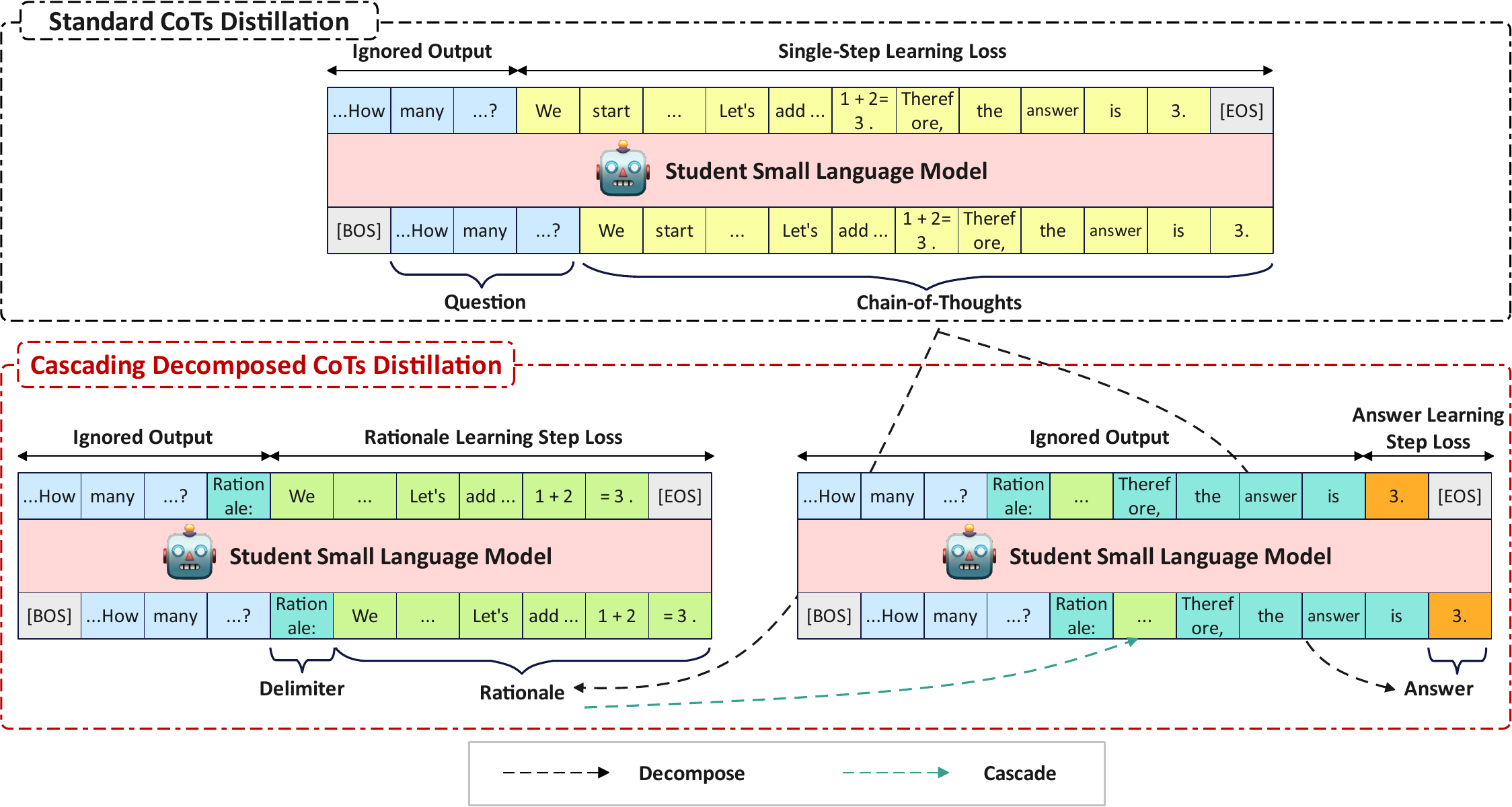}
    \caption{Overview of our proposed method \textbf{Cascading Decomposed CoTs Distillation (CasCoD)}. Different from the standard CoTs distillation, we decompose the single CoT learning step into two comprehensive learning steps including the rationale learning step and the answer learning step, and then learn them in a cascaded way.}
    \label{fig:overview}
\end{figure*}
\paragraph{Distilling Knowledge from LLMs} Numerous studies \cite{alpaca, vicuna2023, intruct-tune-with-gpt4} have explored the knowledge distillation from advanced, proprietary LLMs like ChatGPT \cite{openai2023chatgpt}. These efforts typically concentrate on distilling a broad range of abilities via instruction tuning on extensive and varied datasets \cite{wizardllm, laminilm, lion, acl2024-llmdistill}. However, our work is aimed at distilling the CoT reasoning capabilities from LLMs, aligning with the objectives of \citet{acl-teach-slm-reason, acl-lm-are-reasoning-teachers}, who concurrently propose the standard CoTs distillation method that directly fine-tunes SLMs on CoTs extracted from teacher LLMs. \citet{specializing-slm} expands on this by fine-tuning with various reasoning data formats for specializing domain-specific SLMs. \citet{eacl2024-cot-distill} use different teachers, including in-family and out-family, to generate more CoTs for fine-tuning students. \citet{tailored-learning-slm} distill SLMs via learning from self-reflection and feedback in an interactive, multi-round paradigm with teacher LLMs. \citet{acl-findings-distill-step-by-step} propose to learn the rationale and answers as separate goals for optimizing. \citet{explanations-make-better} propose learning the entire CoTs and the single answers to enhance the reasoning of student SLMs. Based on these foundations, \citet{mind-s-mirror} introduce an additional distillation objective, self-evaluation, aiming for SLMs to assess the accuracy of their CoTs akin to LLMs' evaluative processes.
However, these methods are still affected by the spurious correlations due to their optimization objectives. In contrast, we propose a cascading decomposed CoTs distillation approach that reorganizes training objectives to mitigate this issue.

\section{Methodology}
In this section, we introduce our new distillation method that decomposes the single-step leanring process of standard CoTs distillation into two cascaded learning steps, as illustrated in Figure \ref{fig:overview}. 
Formally, the standard CoTs distillation objective $q \rightarrow CoT$ is split into two learning processes, rationale step learning with the objective $q \rightarrow r$ and answer step learning with the objective $q, r \rightarrow a$. Below we first describe how to extract CoTs from teacher LLMs in \S \ref{extract-cot}. Then we describe the standard CoTs distillation method and discuss its limitations in \S \ref{vanilla-cotd}. Finally, we provide a detailed presentation of our method in \S \ref{method-CasCoD}.

\subsection{Extract CoTs From Teacher LLMs}
\label{extract-cot}
The initial phase of the distillation is to extract CoTs from teacher LLMs for each question-answer pair $\{q,a\}$ in a raw dataset. This involves employing the CoT prompting technique \cite{chain-of-thought-prompting}, which guides the teacher LLMs to generate CoTs that follow a prescribed format with multiple reasoning steps. The prompt template is shown in Appendix \ref{appendix:gen CoTs}. Note that CoTs produced by LLMs may not always be correct. To maintain CoTs quality, following the previous work \cite{acl-teach-slm-reason, acl-findings-distill-step-by-step}, we retain only those that match the ground truth answer in the dataset, building a CoT dataset $\mathcal{D}=\{q, CoT\}$ for training the student model. Additionally, to facilitate the introduction of CasCoD, we explicitly split the extracted CoTs into two parts based on predefined rules in CoT prompting, formalizing this as $CoT=r \oplus a$. For instance, we use the phrase "Therefore, the answer is" to divide the CoT, categorizing the text before this delimiter as the rationale $r$ and the text after it as the answer $a$.

\subsection{Preliminaries for CoTs Distillation}
\label{vanilla-cotd}
Previous standard CoTs distillation \cite{acl-teach-slm-reason, acl-lm-are-reasoning-teachers}, referred to as single-step learning, is to teach SLMs to generate the CoT in one time as follows:
\begin{equation}\label{eq:normal}
    \mathcal{L}_{\text{Std-CoT}} = \mathbb{E}_{q, CoT \sim \mathcal{D}}\left[\ell\left(q, CoT\right)\right]
\end{equation}
where $\ell$ signifies the negative log-likelihood loss function, expressed as:
\begin{equation}\label{eq:l}
    \ell\left(x, y\right) = -\sum_{y_t\in y}\log P\left(y_t\mid x,y_{<t}\right)
\end{equation}

However, this method requires the model to simultaneously learn both rationales and answers in a single step. In such a learning setup, question-answer spurious correlations in widespread implicit reasoning datasets \cite{spurious-between-qa-Blodgett} can be readily captured by the model. These correlations degrade the quality of CoT generation during inference, resulting in weak reasoning generalization. In other words, this implicit learning of correlations might lead student models to preset answers after reading the questions, potentially causing a state reduction in the token generation space when producing CoTs.

\subsection{Cascading Decomposed CoTs Distillation}
\label{method-CasCoD}
Different from the training strategy in standard CoTs distillation, our method decomposes its single-step learning process into two cascaded learning steps, one for the rationale learning step and the other for the answer learning step.

For the rationale learning step, each question is combined with a rationale learning delimiter "Rationale:" as the input $q$, with the rationale $r$ produced by the teacher serving as the label for distilling the rationale. With the answer objective removed, this training strategy allows models to engage in learning rationales without the interference of spurious correlations. The loss function of rationale step learning is as follows:
\begin{equation}\label{eq:rationale learning}
    \mathcal{L}_{\text{rationale}}=\mathbb{E}_{q, r, a \sim \mathcal{D}}\left[\ell\left(q,r\right)\right]
\end{equation}

For the answer learning step, we concatenate both the input and output of the rationale learning step with an answer learning delimiter "Therefore, the answer is" as the input, and the answer $a$ serves as the label for distilling the answer. This strategy helps students learn to reason consistently from the question-rationale pair rather than merely presetting spurious answers based solely on the question. The loss function of answer learning step is thus:
\begin{equation}\label{eq:answer learning}
    \mathcal{L}_{\text{answer}}=\mathbb{E}_{q, r, a \sim \mathcal{D}}\left[\ell\left(q\oplus r, a\right)\right]
\end{equation}

Due to the inherent tight connection between rationale learning and answer learning, for each instance in the dataset, we optimize both learning objectives simultaneously for the CoTs distillation:
\begin{equation}\label{eq:train_std_cot}
    \mathcal{L}_{\text{CasCoD}}=(1 - \alpha)\mathcal{L}_{\text{rationale}} + \alpha \mathcal{L}_{\text{answer}}
\end{equation}
where $\alpha$ is a hyperparameter used to weight the loss in the two learning steps.

During inference, student models follow the same pipeline as in training: first, generate a rationale based on the question, and then predict the final answer using the question-rationale pair. The cascading training objectives reduce the probability of student models capturing spurious correlations between questions and answers in the training phase, thereby alleviating potential reasoning biases caused by spurious correlations in the inference stage, thus enhancing CoTs generalizability.

\section{Experiments}
In this section, we conduct extensive experiments and comprehensive analysis to evaluate the effectiveness of our method across both in-domain (IND) and out-of-domain (OOD) datasets.
\subsection{Datasets}
\subsubsection{In-domain}
\paragraph{BIG-Bench Hard (BBH)} \cite{bbh} comprises 27 challenging tasks covering arithmetic, symbolic reasoning et al. from BIG-Bench (BB) \cite{bb}. The majority of the data involve multiple-choice questions, with a few being open-ended. To underscore the superiority of our approach, we chose to perform distillation on this most challenging dataset. Specifically, we randomly divide the BBH dataset into a training set (BBH-train) for distillation and a test set (BBH-test) as the IND evaluation task, in a 4:1 ratio.

\subsubsection{Out-of-domain}
\paragraph{BIG-Bench Sub (BB-sub).} BB is a popular benchmark consisting of 203 tasks covering a wide range of topics, including mathematics, common-sense reasoning, and various other domains. For ease of evaluation, we filter the subtasks within BB based on subtask keywords, specifically focusing on tasks related to "multiple-choice" and "reasoning"\footnote{For detailed descriptions of the subtasks in BIG-Bench, please refer to \url{https://github.com/google/BIG-bench/blob/main/bigbench/benchmark_tasks/README.md}.}, and ensure that tasks from BBH were excluded, resulting in 61 subtasks.
Then we randomly sample up to 100 instances for each subtask, resulting in the creation of BB-sub.

\paragraph{AGIEval} \cite{agieval} is a renowned human-centric benchmark used to assess LMs' reasoning abilities, whose tasks span various domains, including college entrance exams (English / Math / Law), logic tests et al. We evaluate our method on the subtasks that are related to multiple-choice questions in the English language.

\paragraph{AI2 Reasoning Challenge (ARC)} \cite{arc} consists of ARC-Easy (ARC-E) and ARC-Challenge (ARC-C). The distinction lies in ARC-E consisting of relatively simpler questions from middle and high school science exams, while ARC-C comprises more complex and challenging questions. We utilize the testing set of the ARC dataset for evaluation. The statistics of the all above datasets can be found in Appendix \ref{appendix:data-stat}.

\subsection{Models \& Baselines \& Setup}
\paragraph{Models} We employ the contemporary, popular open-source language model LLaMA2-7B \cite{llama2} as the student SLM. Considering the pricing and capabilities, we utilize OpenAI's powerful black-box LLM, ChatGPT\footnote{\url{https://chat.openai.com/chat}. We utilize the $gpt-3.5-turbo-0613$ for CoTs extraction.}, as the teacher. We query ChatGPT to annotate the CoT data with the same manual prompt used in the previous work \cite{bbh}. 

\paragraph{Baselines} We compare our method with the following baselines: (1) \textbf{Teacher \& Vanilla Student} under various settings, e.g., Zero-shot (+CoT) or Few-shot (+CoT), for showing the impact of distilling reasoning ability from LLMs. (2) \textbf{Std-CoT} \cite{acl-teach-slm-reason, acl-lm-are-reasoning-teachers}, which is the standard CoTs distillation method that directly fine-tune student models on the CoTs data. (3) \textbf{Step-by-step} \cite{acl-findings-distill-step-by-step} is a multi-task CoTs distillation method that distills rationales and answers separately. (4) \textbf{MT-CoT} \cite{explanations-make-better} is also a multi-task CoTs distillation method, but unlike Step-by-step, it simultaneously optimizes the objectives of answer prediction and entire CoTs learning. (5) \textbf{SCOTT} that enhances the reasoning consistency of the student model by introducing additional counterfactual data.

\paragraph{Setup}
We employ LoRA \cite{lora} for parameter-efficient fine-tuning of the student SLMs. In \S \ref{impact-of-weight}, our empirical results indicate that the optimal weight is set $\alpha$ at 0.3. However, to mitigate the effects of unbalanced weighting, we include an additional method setup for comparison against the baselines in Table \ref{tab:main-result}, labeled CasCoD ($\alpha$ = 0.5). All experiments are conducted using a mixed-precision training strategy on 4 $\times$ A100 GPUs. For the inference stage, vLLM\footnote{https://github.com/vllm-project/vllm} \cite{vllm} is utilized to accelerate inference, employing a greedy decoding strategy to generate text on one single A100 GPU. More details on training and hyperparameters can be found in Appendix \ref{appendix:hyperparameter}.
\begin{table*}[htbp]
\resizebox{\linewidth}{!}{
\begin{tabular}{lccccccc|c}
\toprule
\textbf{Method}    & \textbf{Distill?} & \textbf{Gen CoT?}  & \textbf{BBH-test} & \textbf{BB-sub} & \textbf{AGIEval} & \textbf{ARC-E} & \textbf{\hspace{0.5em}ARC-C\hspace{0.5em}} & \multirow{2}{*}{\textbf{\hspace{0.5em}AVG\hspace{0.5em}}} \\ \cmidrule{1-8}
In-domain?          & \multicolumn{1}{|c}{} & \multicolumn{1}{c|}{}  & \checkmark                 & \ding{53}                     & \ding{53}                & \ding{53}              & \ding{53}              &                               \\ \midrule
\multicolumn{9}{c}{\textbf{Teacher: ChatGPT (gpt-3.5-turbo)}}                                                                                                                                      \\ \midrule
Zero-shot-CoT & \multicolumn{1}{|c}{\ding{53}}& \multicolumn{1}{c|}{\checkmark} & 42.7              & 44.1                  & 49.5             & 91.9           & 81.1           & 61.9\\
 Few-shot-CoT& \multicolumn{1}{|c}{\ding{53}}& \multicolumn{1}{c|}{\checkmark} & 73.1& -& -& -& -&-\\ \midrule
 \multicolumn{9}{c}{\textbf{Student: LLaMA2-7B}}                                                                                                                                      \\ \midrule
Zero-shot     & \multicolumn{1}{|c}{\ding{53}}& \multicolumn{1}{c|}{\ding{53}} & 14.8              & 15.5                  & 6.9              & 18.2           & 13.9           & 13.9\\
Zero-shot-CoT & \multicolumn{1}{|c}{\ding{53}}& \multicolumn{1}{c|}{\checkmark} & 10.6              & 7.7                   & 7.1              & 18.4           & 14.8           & 11.7\\
Few-shot     & \multicolumn{1}{|c}{\ding{53}}& \multicolumn{1}{c|}{\ding{53}} & 15.1              & 28.5                  & 25.5             & 25.5           & 25.4           & 24.0\\
Few-shot-CoT & \multicolumn{1}{|c}{\ding{53}}& \multicolumn{1}{c|}{\checkmark} & 16.3              & 25.3                  & 9.9              & 17.2           & 17.2           & 17.2\\
Answer-SFT & \multicolumn{1}{|c}{\ding{53}}& \multicolumn{1}{c|}{\ding{53}} & 51.5             & 33.2                 & 31.2             & 71.6          & 53.7           & 48.2\\ \midrule
Std-CoT \cite{acl-teach-slm-reason}       & \multicolumn{1}{|c}{\checkmark}& \multicolumn{1}{c|}{\checkmark} & 54.2              & 28.7                  & 21.6             & 59.6           & 45.1           & 41.8\\
SCOTT \cite{scott}             & \multicolumn{1}{|c}{\checkmark}& \multicolumn{1}{c|}{\checkmark} & 42.4              & 18.8                  & 13.0& 45.7           & 34.1           & 30.8                          \\
MT-CoT \cite{explanations-make-better}            & \multicolumn{1}{|c}{\checkmark} & \multicolumn{1}{c|}{\checkmark}& \underline{56.8} & 30.3& 22.0& 49.4           & 38.2           & 39.3\\
Step-by-step \cite{acl-findings-distill-step-by-step}      & \multicolumn{1}{|c}{\checkmark} & \multicolumn{1}{c|}{\checkmark}& 42.4              & 27.7                  & \textbf{28.8}    & 68.5& 48.6           & 43.2                          \\
CasCoD (ours, $\alpha=0.5$)        & \multicolumn{1}{|c}{\checkmark}& \multicolumn{1}{c|}{\checkmark} & 52.5     & \underline{36.4}          & 28.1 & \textbf{71.8} & \textbf{54.7} & \underline{48.7}           \\
CasCoD* (ours, $\alpha=0.3$)        & \multicolumn{1}{|c}{\checkmark}& \multicolumn{1}{c|}{\checkmark} & \textbf{59.4}     & \textbf{37.0}           & \underline{28.3} & \underline{70.6} & \underline{52.7} & \textbf{49.6}              \\\bottomrule
\end{tabular}
}
\caption{Accuracy (\%) on in-domain and out-of-domain datasets with different methods. We employ "Let's think step by step" \cite{llm-zero-shot-reasoners} for Zero-shot-CoT settings and the manually curated prompt \cite{bbh} for Few-shot-CoT settings. The best performance among distilled student models is marked in \textbf{bold}, and the second-best performance is indicated by an \underline{underline}.}
\label{tab:main-result}
\end{table*}
\subsection{Main Results}
\label{ex:main results}
Table \ref{tab:main-result} presents the automatic evaluation results of our proposed CasCoD and baselines on both IND and OOD benchmark reasoning datasets. 

\paragraph{CoTs distillation enhances the reasoning performance of students.} Comparing with the Zero-shot-CoT and Few-shot-CoT settings of student models, the performance of those with distillation is significantly improved by learning the teacher LLM's CoTs. Except for BB-sub, the student model has 3-4 times improvement compared to vanilla ones across all datasets.

\paragraph{CasCoD overcomes limitations of distillation baselines in OOD performance.} From the Table \ref{tab:main-result}, we can find that Answer-SFT on the OOD datasets outperforms all the distillation baselines by an average of 5\%, which indicates that it seems student models' performance decreases when learning the CoTs. This pattern is also noticeable in models without distillation, as evidenced by the comparison between Zero-shot and Zero-shot-CoT (or Few-shot and Few-shot-CoT) settings. We attribute this to spurious correlations between questions and answers as introduced in Figure \ref{fig:illustration} (b), which students can easily learn. The distillation baselines that require students to consider predicting answers while generating the rationale, inadvertently make the simpler task of answer prediction interfere with the rationale learning, thus reducing the generalization of CoTs. In contrast, CasCoD* not only surpasses Answer-SFT by 7.9\% in IND datasets but also achieves comparable results in OOD scenarios. This underscores the effectiveness of our cascade two step learning strategy,  which restructures training objectives to mitigate the impact of spurious correlations, in enhancing reasoning capabilities across diverse datasets.
\paragraph{CasCoD significantly outperforms the distillation baselines across IND and OOD datasets.} From Table \ref{tab:main-result}, it can be observed that CasCoD significantly outperforms baselines on both IND and OOD datasets in most cases, regardless of whether the loss is weighted. Specifically, CasCoD* secures an average in-domain improvement of 5.2\% and an out-of-domain enhancement of 8.4\% over the Std-CoT, along with an overall 6.4\% improvement compared to the multi-task learning (Step-by-step) approach. Impressively, CasCoD* achieves 80.1\% of the teacher LLM's performance in Zero-shot-CoT settings. These results underscore the efficacy of CasCoD, significantly boosting the generative capabilities of CoTs on unseen tasks.
\begin{figure}[htb]
	\centering
	\includegraphics[width=\linewidth]{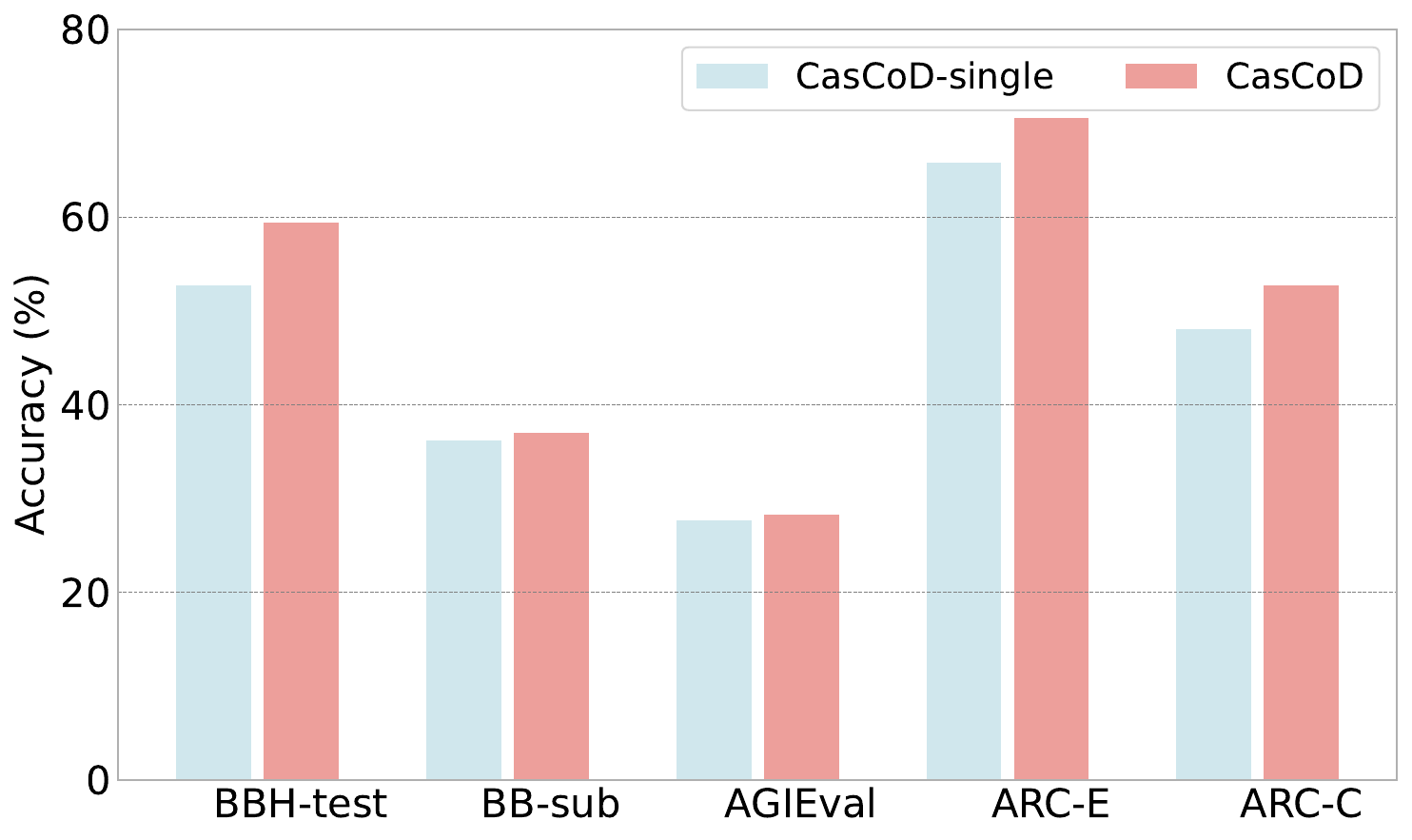}
	\caption{Comparison between two-step and single-step training implementations of CasCoD.}
	\label{ablation-on-multi-steps}
\end{figure}
\begin{figure*}[hbt]
	\centering
	\includegraphics[width=\linewidth]{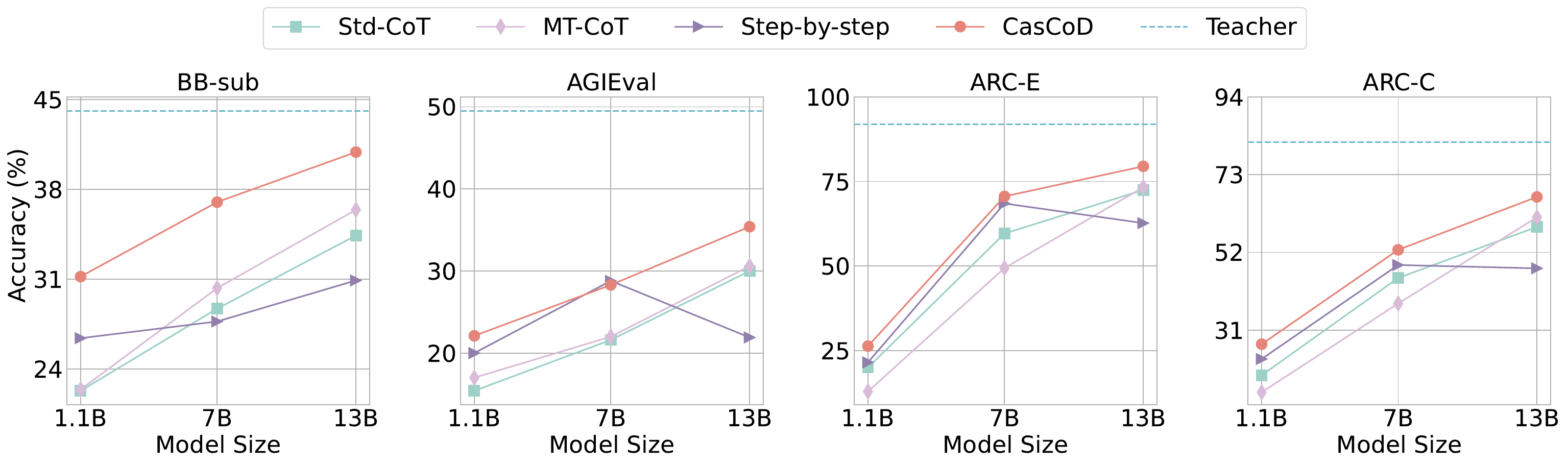}
	\centering
	\caption{Ablation study on model size for four OOD datasets. The dotted line indicates the performance of the teacher LLM under the Zero-shot-CoT setting. The results in IND dataset can be found in Appendix \ref{appendix:ablation-model-size-ind}.}
	\label{ablation-on-model-size-ood}
\end{figure*}
\begin{figure*}[hbt]
	\centering
	\includegraphics[width=\linewidth]{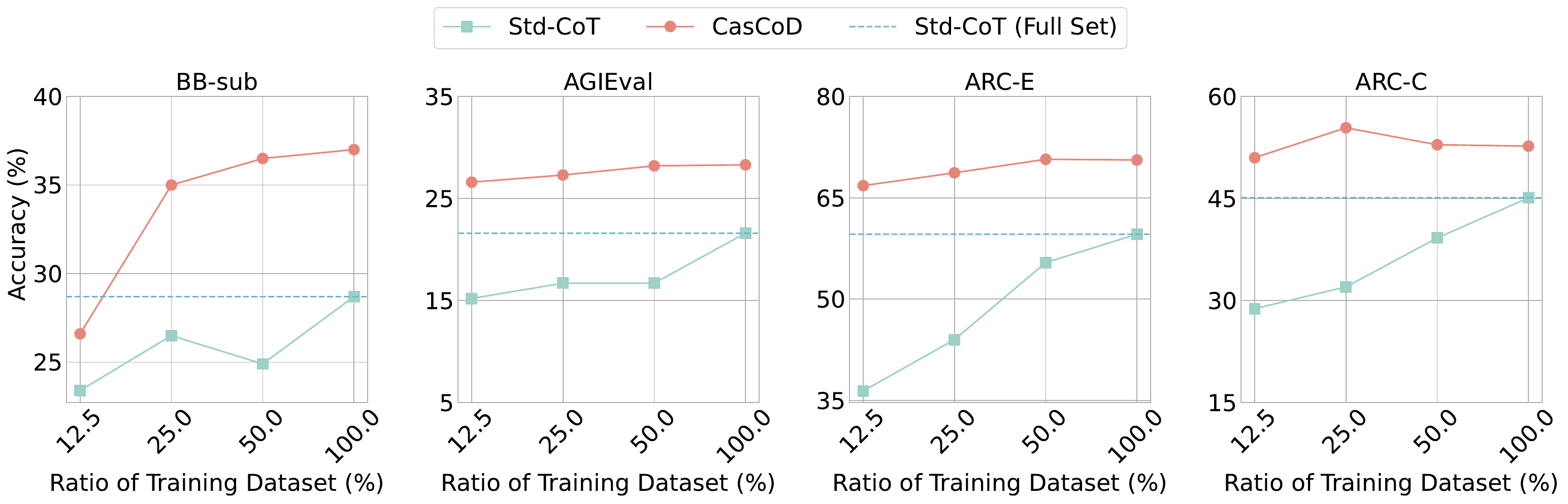}
	\centering
	\caption{Ablation study on training data size for four OOD datasets. The dotted line indicates the performance of fine-tuning the student models by standard CoTs distillation using the full set (100\% of) BBH-train dataset. The results in IND dataset can be found in Appendix \ref{appendix:ablation-data-size-ind}.}
	\label{ablation-on-train-data-size-ood}
\end{figure*}
\subsection{Ablation Study on Model \& Data Sizes}
\label{ablation study}
\paragraph{CasCoD is universally applicable to models of varying sizes.} We perform model distillation on TinyLLaMA-1.1B\footnote{\url{https://huggingface.co/TinyLlama/TinyLlama-1.1B-intermediate-step-1431k-3T}} \cite{tinyllama}, LLaMA2-7B, and LLaMA2-13B, respectively and compare with standard CoTs distillation (Std-CoT) and multi-task distillation (MT-CoT \& Step-by-step). In Figure \ref{ablation-on-model-size-ood} and \ref{ablation-on-model-size-ind}, we can find that CasCoD consistently outperforms the baselines on both IND and OOD datasets across various sizes of student models. Notably, the performance improvement of our method is the most obvious in the BB-sub, where the performance of the 13B student model reaches 92.7\% of the teacher LLM's performance.
Furthermore, as model sizes increase, the performance gap between CasCoD and the baselines widens on OOD datasets, highlighting CasCoD's superior efficiency in distilling CoTs for larger models.
\paragraph{CasCoD is universally applicable to models of different architectures.} We perform model distillation on CodeLLaMA-7B \cite{llama2}, LLaMA3-8B \cite{llama3} and Mistral-7B-v0.2 \cite{mistral}, respectively, and compare with Std-CoT and Step-by-step. From the Table \ref{tab:different model arcs}, we can see that regardless of whether it's CodeLLaMA, LLaMA3, or Mistral, CasCoD significantly outperforms the baselines on OOD tasks, demonstrating its high effectiveness and scalability. Particularly, on the powerful base model Mistral, the superiority of our method is further amplified.
\paragraph{CasCoD significantly outperforms standard CoTs distillation on OOD with much less training data.} In Figure \ref{ablation-on-train-data-size-ood}, CasCoD achieves a 6.3\% improvement over Std-CoT on the BB-sub dataset, using only 25\% of the full BBH-train data. In the case of other OOD datasets, CasCoD requires merely 12.5\% of the full training data to surpass the Std-CoT trained with the full dataset by 5\% to 7\% in performance. These results demonstrate the efficiency of CasCoD, capable of enhancing CoTs generalization with a smaller amount of CoTs data.
\begin{table*}[htbp]
\small
\centering
\resizebox{\linewidth}{!}{
\begin{tabular}{lcccccc}
\toprule
\textbf{Models \& Methods} & \textbf{BBH-test} & \textbf{BB-sub} & \textbf{AGIEval} & \textbf{ARC-E} & \textbf{ARC-C} & \textbf{AVG} \\
\midrule
\multirow{1}{*}{In-domain?}& \checkmark & \ding{53} & \ding{53} & \ding{53} & \ding{53} &  \\
\midrule
CodeLLaMA-7B + Std-CoT & \textbf{56.2}& \underline{29.7}& 19.2& \underline{42.0}& \underline{32.2}& \underline{35.9}\\
CodeLLaMA-7B + Step-by-step & 40.7& 29.0& \underline{23.9}& 41.5& \textbf{32.8}& 33.6\\
CodeLLaMA-7B + CasCoD & \underline{54.8}& \textbf{35.4}& \textbf{25.8}& \textbf{42.9}& 31.7& \textbf{38.1}\\\midrule
LLaMA3-8B + Std-CoT & \textbf{66.9}& 33.9& 32.7& 69.8& 60.2& 52.7\\
LLaMA3-8B + Step-by-step & 44.2& \underline{35.5}& \underline{38.8}& \underline{83.7}& \underline{70.7}& \underline{54.5}\\
LLaMA3-8B + CasCoD & \underline{65.2}& \textbf{42.9}& \textbf{40.1}& \textbf{87.2}& \textbf{74.0}& \textbf{61.9}\\\midrule
Mistral-7B-v0.2 + Std-CoT & \textbf{72.2}& 37.6& \underline{32.0}& 68.8& 57.9& \underline{53.7}\\
Mistral-7B-v0.2 + Step-by-step & 56.4& \underline{38.9}& 20.1& \underline{76.4}& \underline{62.3}& 50.8\\ 
Mistral-7B-v0.2 + CasCoD & \underline{71.7}& \textbf{42.5}& \textbf{40.1}& \textbf{83.9}& \textbf{74.2}& \textbf{62.5}\\
\bottomrule
\end{tabular}
}
\caption{Accuracy (\%) on IND and OOD datasets with different student models distilled by different methods.}
\label{tab:different model arcs}
\end{table*}
\section{Analysis}
\subsection{Two-Step vs. Single-Step Implementation}
In this subsection, we explore whether CasCoD's two-step training objectives can be achieved in a single-step computation. Upon analysis of the two cascaded learning steps, we find that under teacher-forcing \cite{teacher-forcing}, CasCoD closely mirrors Std-CoT, with key distinctions including adjustable token-level weights and the omission of delimiters in loss calculations. Each sample in CasCoD's original framework undergoes two forward calculation, raising the question of whether a similar outcome is possible with only one.
To investigate this, we introduce a variant, CasCoD-single, which is designed to fulfill the two-step training objectives through a single forward computation.
Figure \ref{ablation-on-multi-steps} indicates that the two-step CasCoD consistently surpasses the single-step variant across all datasets. This underscores that a single forward calculation does not suffice to meet CasCoD's training objectives, emphasizing the critical importance of the cascading two-step learning process.
\begin{figure*}[!htb]
	\centering
	\includegraphics[width=\linewidth]{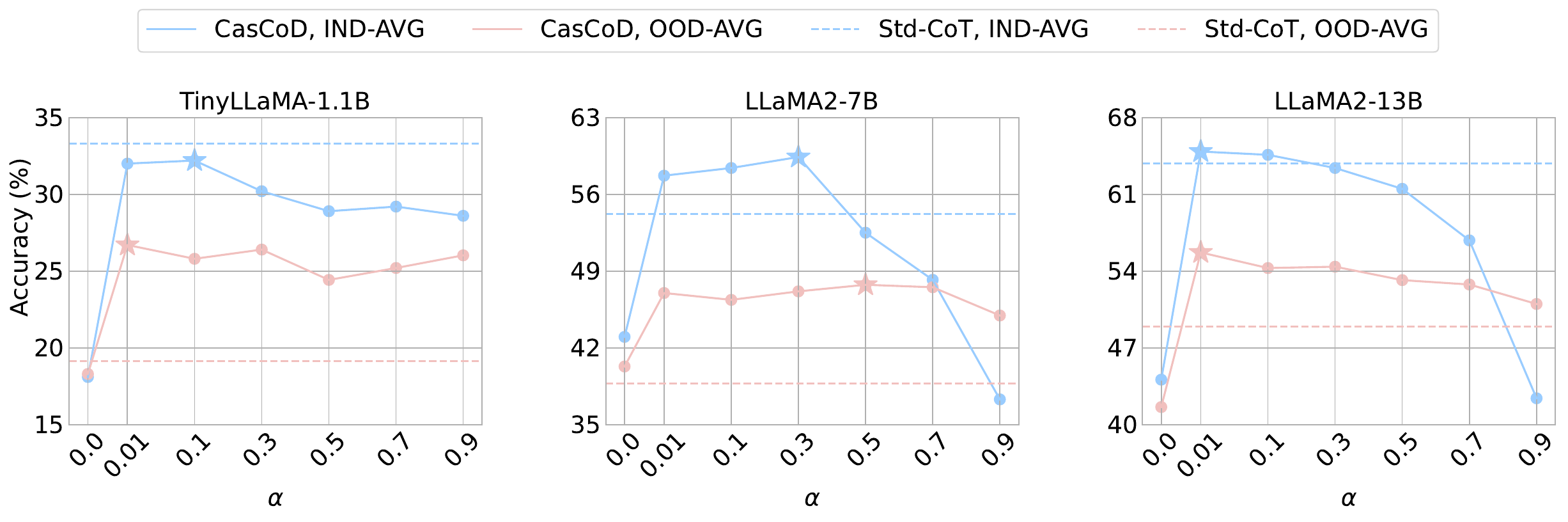}
	\centering
	\caption{Ablation study on task weights $\alpha$. The results are reported by \bluehighlight{IND-AVG} and \redhighlight{OOD-AVG} that respectively denote averge accuracy on IND and OOD datasets. The best performance among weights are marked with "\ding{73}".}
	\label{ablation-on-weights}
\end{figure*}
\subsection{Impact of Weights}
\label{impact-of-weight}
In this subsection, we explore how variations in weights affect the performance of models with different parameter sizes on both IND and OOD datasets, as shown in Figure \ref{ablation-on-weights}. 
\paragraph{Students' performance is not sensitive to weights on OOD datasets.} From the figure, we observe that regardless of weight changes, CasCoD consistently outperforms Std-CoT in OOD by average, even at $\alpha=0.9$ (meaning the model allocated only 10\% of its attention to rationales generation). This demonstrates that CasCoD exhibits robust generalization in OOD and also underscores the effectiveness of decomposing CoTs for distillation.
\paragraph{CasCoD is more robust for smaller student models.} We observe that the 1.1B model shows less variation in performance compared to the 7B and 13B models in IND. Notably, the performance of the 13B model drops sharply as $\alpha$ changes from 0.5 to 0.9, indicating that larger models are more susceptible to weight adjustments in the IND dataset.
\paragraph{Prioritizing the rationale over the answer yields better results.} It is evident that across different model sizes, the optimal weights on both IND and OOD datasets range approximately from 0.01 to 0.3, indicating that focusing on the rationale help improve the generalizability of CoTs.
\subsection{Faithfulness of Students}
To ensure that the rationale provided by students supports their predicted answers, another metric for evaluating CoTs distillation is the faithfulness of students.
\begin{table}[htbp]
\footnotesize
\centering
\begin{tabular}{c|cc|c}
\toprule
\textbf{Method}       & \textbf{ChatGPT}       & \textbf{GPT4}          & \textbf{AVG}           \\ \midrule
Teacher      & 41.1          & 36.3          & 38.7          \\ \midrule
Std-CoT      & 40.8          & 29.8          & 35.3          \\
SCOTT        & 36.2          & 29.4          & 32.8          \\
MT-CoT       & 36.2          & 25.8          & 31            \\
Step-by-step & 6.6           & -0.1          & 3.25          \\
CasCoD (ours)  & \textbf{40.8} & \textbf{31.6} & \textbf{36.2} \\ \bottomrule
\end{tabular}
\caption{Faithfulness (LAS, \%) of the compared methods with different LLM evaluators on the IND dataset. The prompt templates can be found in Appendix \ref{appendix:LAS}}.
\label{tab:faithfulness}
\end{table}
Following the previous work \cite{scott}, we use the LAS metric \cite{LAS}, whose core idea is to measure the extent that the rationales $r'$ aid a simulator in predicting the answers $a'$, defined as:
\begin{equation}\label{eq:LAS}
    LAS=Acc\left(q, r' \rightarrow a'\right) - Acc\left(q \rightarrow a'\right)
\end{equation}
where we employ ChatGPT and GPT4 as the simulator, respectively. The results are shown in Table \ref{tab:faithfulness}.
CasCoD is observed to generate rationales that are more consistent with answers than baselines. This suggests that despite CasCoD's multi-step learning process, the introduction of cascading learning ensures that students can faithfully reason.

\subsection{Hypothesis Validation}
In this subsection, we aim to validate our hypothesis that student learning of spurious correlations affects the quality of their generated rationales. Due to the nature of causal language models---autoregression, we incorporate "The answer is" into input prompts to directly prompt models to provide the final answer immediately (which we treat as the "preset answer" referred to \S\ref{intro}) after reading the question, rather than generating rationales before providing an answer. We define the following metrics:
\begin{enumerate}
    \item \textbf{a}: The proportion of incorrect CoT reasoning when the preset answer is wrong. A higher ratio indicates a greater negative impact of the preset answer on CoT reasoning.
    \item \textbf{b}: The proportion of correct CoT reasoning when the preset answer is correct. A higher ratio indicates a greater positive impact of the preset answer.
    \item \textbf{c} = a - b (combine a and b).
\end{enumerate}
We compare CasCoD with Std-CoT using these metrics on four OOD tasks. The results are shown in Table \ref{tab:hypothesis validation}. We observe that Std-CoT significantly outperforms CasCoD in metrics 'a' and 'c', while significantly underperforms CasCoD in metric 'b' on all four OOD tasks. This indicates that our method can selectively utilize spurious correlations to some extent, suppressing the negative effects of incorrect preset answers on reasoning and reinforcing the positive effects of correct preset answers on reasoning, thereby enhancing performance on OOD tasks, which experimentally validate our hypothesis.
\begin{table}[htbp]
\scriptsize
\centering
\begin{tabular}{c|cccc}
\toprule
\textbf{Method \& Metric}& \textbf{BB-sub}& \textbf{AGIEval}& \textbf{ARC-E} &\textbf{ARC-C}\\ \midrule
Std-CoT (a) $\downarrow$& 79.7& 82.2& 63.7&50.8\\
Std-CoT (b) $\uparrow$& 47.5& 30.6& 58.1&69.7\\
\cellcolor{gray!30} Std-CoT (c) $\downarrow$&\cellcolor{gray!30} 32.2&\cellcolor{gray!30} 51.6&\cellcolor{gray!30} 5.6&\cellcolor{gray!30}-18.9\\
CasCoD (a) $\downarrow$& 73.3& 77.9& 61.8&50.9\\
 CasCoD (b) $\uparrow$& 57.4& 36.7& 71.4&83.9\\
\cellcolor{gray!30} CasCoD (c) $\downarrow$&\cellcolor{gray!30} \textbf{15.9}&\cellcolor{gray!30} \textbf{41.2}&\cellcolor{gray!30} \textbf{-9.6}&\cellcolor{gray!30} \textbf{-33.0}\\ \bottomrule
\end{tabular}
\caption{Hypothesis validation results (\%) on four OOD tasks. The hypothesis is better supported when 'a' is lower, 'b' is higher, and 'c' is lower.}.
\label{tab:hypothesis validation}
\end{table}
\subsection{Case Study}
Due to page limitations, we provide a systematic case study in Appendix \ref{appendix:case-study} to illustrate the improvement in CoT generalizability.

\section{Conclusion}
In this paper, we propose a simple yet effective CoTs distillation method CasCoD to address the issue of question-answer spurious correlations that previous CoTs distillation methods suffer from. Specifically, we decompose the traditional single-step learning process into two cascaded learning steps and restructure their training objectives. Extensive experiments show that CasCoD significantly outperforms the baselines across both IND and OOD datasets. Further analysis reveals that CasCoD is robust to model size, training data size, and weights and can lead to faithful student models.
\section*{Limitations}
Considering the cost such as API calls and GPU training expenses, we only choose ChatGPT as the teacher LLM and the widely-used model LLaMA2 as the student SLM. Employing GPT-4 as the teacher provides high-quality CoTs, which could better validate the effectiveness of our proposed method CasCoD.
Besides, when distilling the student model using CasCoD, it requires two forward computations, increasing the training time cost.

Besides, in our study, we explore distilling CoTs into two cascading steps, which is an initial step toward understanding finer decompositions. Research \cite{llm-emergent-mirage} suggests that the emergent abilities of LLMs result from managing multiple sub-tasks simultaneously, hinting at the potential for more intricate cascading steps in CoTs. Our current work does not yet define the precise rules for such more steps decomposition, nor the optimal timing and methods for focusing learning on specific steps. These aspects remain areas for future exploration, intended to refine and extend the CoT reasoning capabilities of SLMs as suggested by our findings.

\section*{Ethics Statement}
Our work utilizes CoT data extracted from ChatGPT for distillation, which may result in inheriting the social biases \cite{llm-social-bias} and hallucination \cite{Hallucinations} present in LLMs. However, we are optimistic that future advancements in resolving these issues in LLMs will naturally lead to the development of student models with reduced toxicity.
\bibliography{anthology,custom}
\bibliographystyle{acl_natbib}

\appendix
\section{Ablation Study on In-domain Dataset}
\subsection{W.R.T. Model Size}
\label{appendix:ablation-model-size-ind}
\begin{figure}[hbt]
	\includegraphics[width=\linewidth]{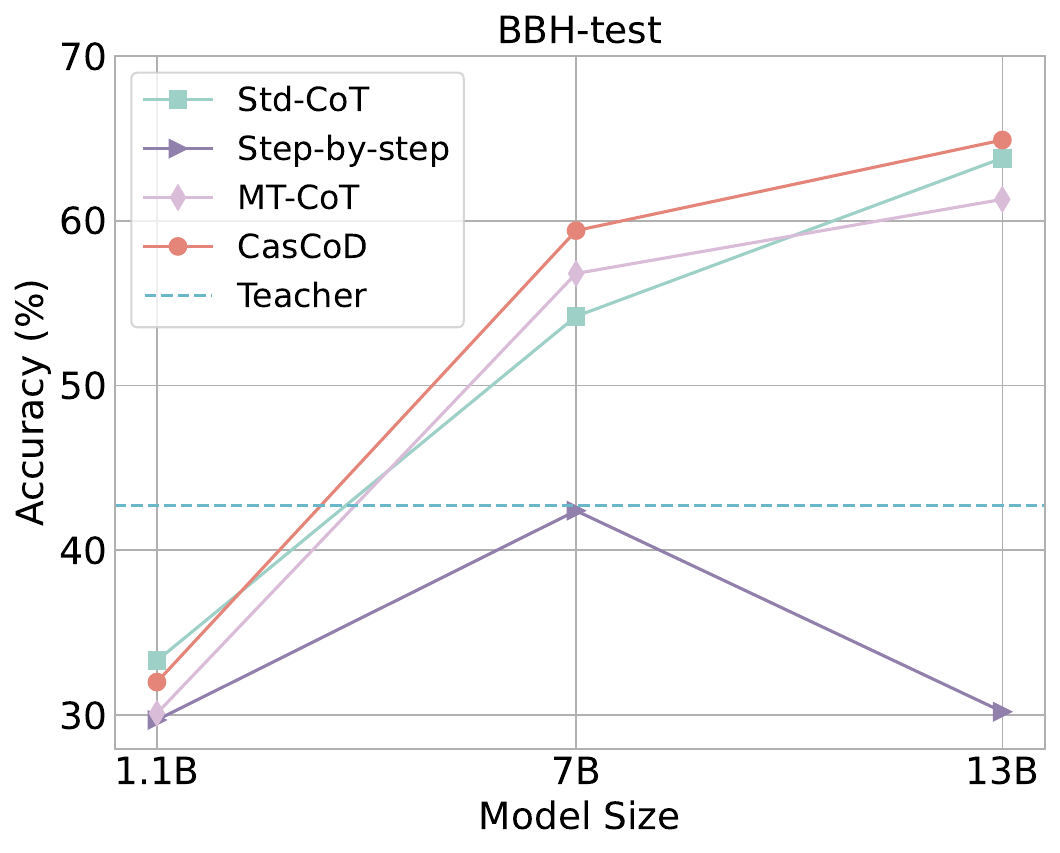}
	\centering
	\caption{Ablation study on model size in the IND (BBH-test). The dotted line indicates the performance of the teacher LLM under the Zero-shot-CoT setting.}
	\label{ablation-on-model-size-ind}
\end{figure}

The results of the model size ablation study on IND datasets are presented in Figure \ref{ablation-on-model-size-ind}. We observe that CasCoD outperforms the baselines on both the 7B and 13B model and significantly surpasses the teacher LLMs in the Zero-shot CoT setting.
\subsection{W.R.T. Training Data Size}
\label{appendix:ablation-data-size-ind}
\begin{figure}[hbt]
	\includegraphics[width=\linewidth]{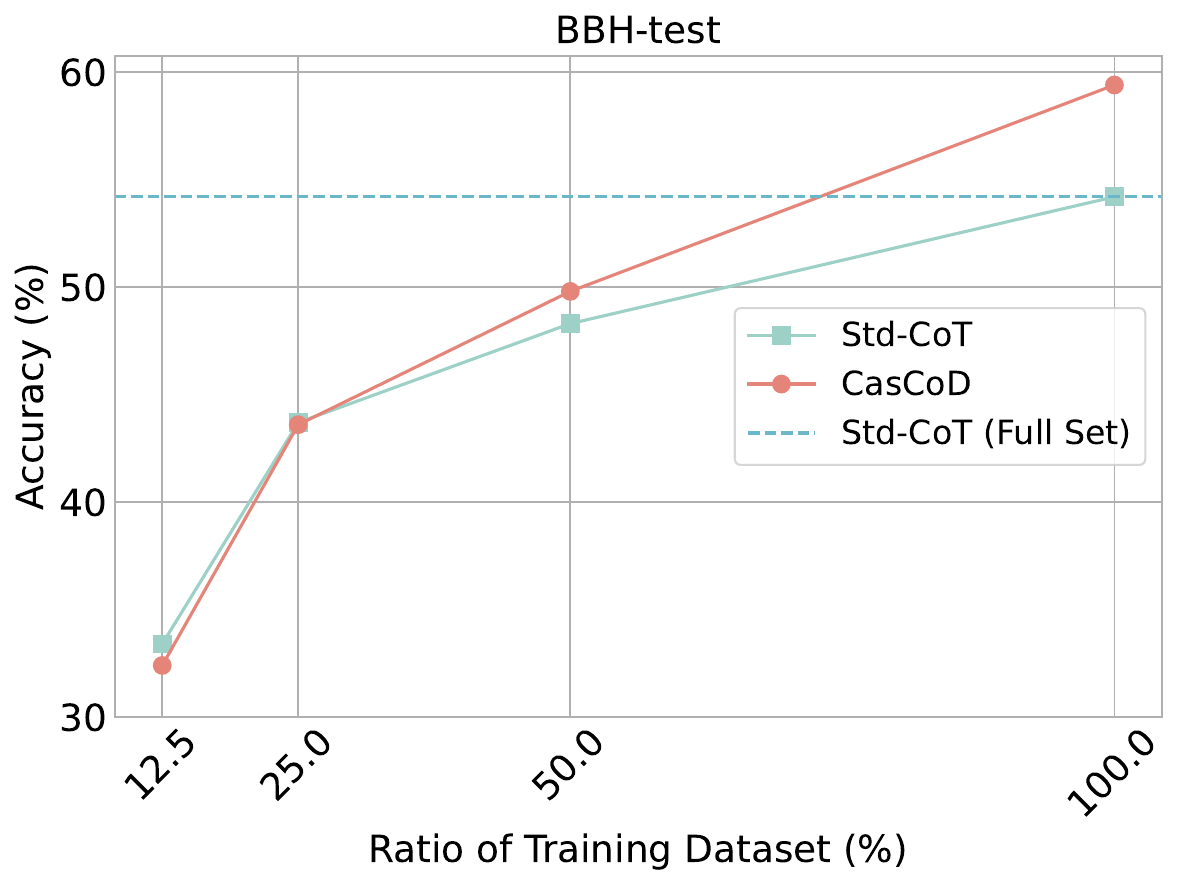}
	\centering
	\caption{Ablation study on training data size in the IND (BBH-test). The dotted line indicates the performance of fine-tuning the student models by standard CoT distillation using the full set (100\% of) BBH-train dataset.}
	\label{ablation-on-train-data-size-ind}
\end{figure}

The results of the training data ablation study on IND datasets, as shown in Figure \ref{ablation-on-train-data-size-ind}, indicate that CasCoD outperforms standard CoTs distillation across various sizes of training data. This demonstrates the efficiency of our proposed method.
\section{Details of Experiment}
\subsection{Dataset Statistics}
\label{appendix:data-stat}
Table \ref{tab:agieval_sat}, Table \ref{tab:arc_sat}, Table \ref{tab:bbheval_sat} and Table \ref{tab:bbsubeval_sat} show the data statistics of AGIEval, ARC, BIG-Bench Hard (BBH) and BIG-Bench Sub (BB-sub), respectively.
\begin{table}[!h]
\footnotesize
\centering
\begin{tabular}{l|l|c|c}
\toprule
 \textbf{No.}&\textbf{Task}                         & \textbf{Size}& \textbf{\# Choices} \\ \midrule
1&AQuA-RAT               & 254                  & 5                   \\
2&LogiQA-EN& 651                  & 4                   \\
3&LSAT-AR                & 230                  & 5                   \\
4&LSAT-LR                & 510                  & 5                   \\
5&LSAT-RC                & 269                  & 5                   \\
6&SAT-Math               & 220                  & 4                   \\
7&SAT-EN& 206                  & 4                   \\
8&SAT-EN (w/o Psg.)& 206                  & 4      \\     
\midrule
 &\textbf{Sum} & 2546& - \\\bottomrule         
\end{tabular}
\caption{Statistics of AGIEval dataset.}
\label{tab:agieval_sat}
\end{table}

\begin{table}[htbp]
\footnotesize
\centering
\begin{tabular}{l|c|c}
\toprule
\textbf{Task}          & \textbf{Size}& \textbf{\# Choices} \\ \midrule
ARC-E               & 2376                  & 4-5                   \\
ARC-C & 1172                  & 4-5                   \\   \bottomrule
\end{tabular}
\caption{Statistics of ARC test dataset.}
\label{tab:arc_sat}
\end{table}

\begin{table}[!htb]
\scriptsize
\centering
\begin{tabular}{l|l|c|c}
\toprule
 \textbf{No.}&\textbf{Task}                         & \textbf{Size}& \textbf{\# Choices} \\ \midrule
 1&Boolean Expressions                   & 250                  & 2                   \\
 2&Causal Judgement                      & 187                  & 2                   \\
 3&Date Understanding                    & 250                  & 6                   \\
 4
&Disambiguation QA                     & 250                  & 4                   \\
  5
&Dyck Languages& 250&-\\
 6
&Formal Fallacies Syllogisms Negation& 250                  & 2                   \\
 
7
&Geometric Shapes                      & 250                  & 11                  \\
 8
&Hyperbaton (Adjective Ordering)& 250                  & 2                   \\
 
9
&Logical Deduction (3 objects)& 250                  & 3\\
 10
&Logical Deduction (5 objects)& 250                  & 5\\
 
11
&Logical Deduction (7 objects)& 250                  & 7\\
 12
&Movie Recommendation                  & 250                  & 5                   \\
  
13
&Multi-Step Arithmetic& 250&-\\
 14
&Navigate                              & 250                  & 2                   \\

15
&Object Counting& 250&-\\
 16
&Penguins in a Table                   & 146                  & 5                   \\

17
&Reasoning about Colored Objects       & 250                  & 18                  \\
 18
&Ruin Names                            & 250                  & 11                   \\

19
&Salient Translation Error Detection   & 250                  & 6                   \\
 20
&Snarks                                & 178                  & 2                   \\
 
21&Sports Understanding                  & 250                  & 2                   \\
 22&Temporal Sequences                    & 250                  & 4                   \\
 
23&Tracking Shuffled Objects (3 objects)& 250                  & 3\\
 24&Tracking Shuffled Objects (5 objects)& 250                  & 5\\
 
25&Tracking Shuffled Objects (7 objects)& 250                  & 7\\
 26&Web of Lies                           & 250                  & 2 \\ 
  
27&Word Sorting& 250&-\\    
\midrule
 &\textbf{Sum} & 6511& - \\\bottomrule             
\end{tabular}
\caption{Statistics of BIG-Bench Hard dataset.}
\label{tab:bbheval_sat}
\end{table}
\begin{table}[!htb]
    \scriptsize
    \centering
    \begin{tabular}{l|l|c|c}
    \toprule
     \textbf{No.}&\textbf{Task}                         & \textbf{Size}& \textbf{\# Choices} \\ \midrule
     1 &abstract\_narrative\_understanding& 100& 5\\
     2 &anachronisms& 100& 2                   \\
     3 &analogical\_similarity& 100& 7\\
     4 &analytic\_entailment& 70& 2\\
     5 & cause\_and\_effect& 100&2\\
      6 &checkmate\_in\_one& 100&26\\
     7 &cifar10\_classification& 100& 10\\
     8 &code\_line\_description& 60& 4\\
     9 &conceptual\_combinations& 100& 4\\
     10 &crass\_ai& 44& 4\\
     11 &elementary\_math\_qa& 100& 5                   \\
     12 &emoji\_movie& 100& 5\\
      13 &empirical\_judgments& 99&3\\
     14&english\_russian\_proverbs& 80& 4\\
     15&entailed\_polarity& 100& 2\\
     16&entailed\_polarity\_hindi& 100& 2\\
     17&epistemic\_reasoning& 100& 2\\
     18&evaluating\_information\_essentiality& 68& 5\\
     19&fantasy\_reasoning& 100& 2                   \\
     20&figure\_of\_speech\_detection& 59& 10\\
     21
    &goal\_step\_wikihow
    & 100& 4\\ 
     22
    &gre\_reading\_comprehension
    & 31& 5\\
     23
    &human\_organs\_senses
    & 42& 4\\ 
     24
    &identify\_math\_theorems
    & 53& 4\\
     25
    &identify\_odd\_metaphor
    & 47& 5\\
     26& implicatures& 100&2\\ 
     27&implicit\_relations
    & 82& 25\\
     28&indic\_cause\_and\_effect
    & 100& 2 \\ 
     29&intersect\_geometry
    & 100& 26\\
     30&kanji\_ascii
    & 100& 5\\ 
     31&kannada
    & 100& 4\\
     32&key\_value\_maps
    & 100& 2 \\ 
     33&logic\_grid\_puzzle
    & 100& 3\\
     34&logical\_args
    & 32& 5\\ 
     35&logical\_fallacy\_detection
    & 100& 2\\
     36&metaphor\_boolean
    & 100& 2\\ 
     37&metaphor\_understanding
    & 100& 4\\ 
     38&minute\_mysteries\_qa
    & 100& 4\\
     39&mnist\_ascii
    & 100& 10\\ 
     40&moral\_permissibility
    & 100& 2\\
     41&movie\_dialog\_same\_or\_different
    & 100& 2 \\ 
     42&nonsense\_words\_grammar
    & 50& 4\\
     43&odd\_one\_out
    & 86& 5\\ 
     44&parsinlu\_qa
    & 100& 4\\
     45&physical\_intuition
    & 81& 4\\ 
     46&play\_dialog\_same\_or\_different
    & 100& 2\\
     47&presuppositions\_as\_nli
    & 100& 3\\ 
     48&riddle\_sense
    & 49& 5\\ 
     49&similarities\_abstraction
    & 76& 4\\
     50&simple\_ethical\_questions
    & 100& 4\\ 
     51&social\_iqa
    & 100& 3\\
     52&strange\_stories
    & 100& 2 \\ 
     53&strategyqa
    & 100& 2\\
     54&swahili\_english\_proverbs
    & 100& 4\\ 
     55&swedish\_to\_german\_proverbs
    & 72& 4\\
     56&symbol\_interpretation
    & 100& 5\\ 
     57&timedial
    & 100& 3\\
     58&undo\_permutation
    & 100& 5\\ 
     59&unit\_interpretation
    & 100& 5\\ 
     60&vitaminc\_fact\_verification
    & 100& 3\\
     61&winowhy& 100& 2 \\
    \midrule
     &\textbf{Sum} & 5384& - \\
     \bottomrule             
    \end{tabular}
    \caption{Statistics of BIG-Bench sub dataset. We filter the original dataset by retrieving tasks with keywords "multiple choice" and randomly sample up to 100 examples per task. Note, the task in BBH will not be involved in BB-sub.}
    \label{tab:bbsubeval_sat}
\end{table}

\subsection{Hyperparameters Settings}
\label{appendix:hyperparameter}
In our study, we ensure consistency in the hyperparameter settings across all baselines, including our proposed CasCoD approach, to maintain the fairness of our comparative analysis. Here, we detail the hyperparameter configurations employed in our experiments.
\paragraph{Training Steps and Batch Size.} The number of training steps is determined based on the size of the training dataset, the batch size, and the number of gradient accumulation steps required. We maintain a consistent batch size across all baselines to eliminate any performance discrepancies that could arise from varying batch sizes.
\paragraph{Learning Rate.} Our exploratory experiments initially focus on the standard CoTs distillation method using the LLaMA-2 model, revealing that while the batch size had minimal impact on performance, the learning rate was a critical factor. We test learning rates of 1e-4, 2e-4, and 3e-4 and observe optimal performance at 2e-4 across Std-CoT and other distillation baselines as well as our CasCoD. Therefore, we set the learning rate to 2e-4 for all methods involved in our study.
\paragraph{Epochs and Evaluation Strategy.} Throughout our training process, we monitor the training loss curve and note that it generally plateaued by the 15th epoch, suggesting that the models have achieved convergence. Accordingly, we set the number of epochs to 15 for 7B models. The process of determining the number of epochs for other model sizes follows a similar pattern. To mitigate the potential risk of overfitting and to ensure that our evaluation reflects the most effective model configuration, we systematically select the checkpoints from the epoch that demonstrate the best performance on the IND task. These checkpoints are then used to evaluate performance on OOD tasks.

Finally, the detailed hyperparameters in training and inference can be found in Table \ref{tab:train-hyperparameters} and Table \ref{tab:infer-hyperparameters}, respectively.
\begin{table}[!htb]
\scriptsize
\centering
\resizebox{\linewidth}{!}{
\begin{tabular}{l|ccc}
\toprule
\textbf{Hyperparameter} &  \textbf{TinyLLaMA-1.1B}&\textbf{LLaMA2-7B}& \textbf{LLaMA2-13B}\\ \midrule
gradient accumulation steps &4 &4&8\\
per device batch size&  16&16& 8\\
learning rate           &  2e-4&2e-4& 2e-4\\
epoches                 &  20&15& 10\\
max length              &  1024             &1024             & 1024              \\
$\beta$ of AdamW&  (0.9,0.999)&(0.9,0.999)& (0.9,0.999)\\
 $\epsilon$ of AdamW& 1e-8& 1e-8&1e-8\\
$\gamma$ of Scheduler&  0.95&0.95& 0.95\\
weight decay            &  0                &0                & 0                 \\
warmup ratio            &  0                &0                & 0                \\ 
rank of LoRA&  64&64& 64\\ 
$\alpha$ of LoRA&  32&32& 32\\
 target modules& q\_proj, v\_proj& q\_proj, v\_proj&q\_proj, v\_proj\\ 
drop out of LoRA&  0.05&0.05& 0.05\\ 
\bottomrule
\end{tabular}
}
\caption{Training hyperparameters.}
\label{tab:train-hyperparameters}
\end{table}
\begin{table}[!htb]
\footnotesize
\centering
\begin{tabular}{l|cc}
\toprule
\textbf{Arguments}&  \textbf{Student}&\textbf{Teacher}\\ \midrule
do sample&  False&True\\
temperature&  -&0.2\\
 top-p& 1.0&1.0\\
top-k&  -&-\\
max new tokens&  1024             &2048\\
\# return sequences&  1&1\\
\bottomrule
\end{tabular}
\caption{Generation configs of students and teachers.}
\label{tab:infer-hyperparameters}
\end{table}
\section{Prompts}
\subsection{Prompts of Generating CoTs for ChatGPT}
\label{appendix:gen CoTs}
We use the prompt template shown in Table \ref{tab:prompt-gen-CoTs} to call the ChatGPT API to generate the CoTs for the BBH-train datasets.
\begin{table*}[htbp]
\small
\centering
\begin{tabular}{l}
\toprule
\texttt{\{Task Description\}. Your response should conclude with the format "Therefore, the answer is".}\\\\
\texttt{Q: \{Task Example Question No.1\}}\\
\texttt{A: Let's think step by step. \{Human-Curated-CoTs No.1\}.}\\\\
\texttt{Q: \{Task Example Question No.2\}}\\
\texttt{A: Let's think step by step. \{Human-Curated-CoTs No.2\}.}\\\\
\texttt{Q: \{Task Example Question No.3\}}\\
\texttt{A: Let's think step by step. \{Human-Curated-CoTs No.3\}.}\\\\
\texttt{Q: \{QUESTION\}}\\
\texttt{A: Let's think step by step.}\\
\bottomrule
\end{tabular}
\caption{Prompt template of gpt-3.5-turbo for generating the CoTs data with 3 shots.}
\label{tab:prompt-gen-CoTs}
\end{table*}

\subsection{Prompts of Simulators}
\label{appendix:LAS}
We use the prompt templates shown in Table \ref{tab:prompt-q-a} and Table \ref{tab:prompt-q-r-a} to call the ChatGPT and GPT4 API to predict the answers given a question or with an additional rationale, respectively.
\begin{table*}[h]
\small
\centering
\begin{tabular}{l|l}
\toprule
system content & \parbox[c]{13cm}{%
\texttt{You are a helpful and precise assistant for following the given instruction.}
}\\
\midrule
user content & \parbox[c]{13cm}{\texttt{[Instruction] \{Please read the question and then give your answer based on the question without any explanations.\}} \\ \\
\texttt{Task Description: \{TASK\_DESCRIPTION\}}\\ \\
\texttt{Question: \{QUESTION\}} \\ \\
\texttt{Your Answer: } \\
}\\
\bottomrule
\end{tabular}
\caption{Prompt template of simulators for predicting the answers when given the question.}
\label{tab:prompt-q-a}
\end{table*}

\begin{table*}[h]
\small
\centering
\begin{tabular}{l|l}
\toprule
system content & \parbox[c]{13cm}{%
\texttt{You are a helpful and precise assistant for following the given instruction.}
}\\
\midrule 
user content & \parbox[c]{13cm}{%
\texttt{[Instruction] \{Please read the question and the rationale, and then give your answer based on the question and the rationale without any explanations.\}} \\ \\
\texttt{Task Description: \{TASK\_DESCRIPTION\}}\\ \\
\texttt{Question: \{QUESTION\}} \\ \\
\texttt{Rationale: \{RATIONALE\}} \\ \\
\texttt{Your Answer: } \\
}\\
\bottomrule
\end{tabular}
\caption{Prompt template of simulators for predicting the answers when given the question and rationale.}
\label{tab:prompt-q-r-a}
\end{table*}

\section{Case Study}
\label{appendix:case-study}
Here we show 4 cases in Table \ref{tab:bbh-boolean-expresion}, \ref{tab:bbh-reasoning-color}, \ref{tab:sat-math-agieval} and \ref{tab:case-arc} to compare the CoT generated by CasCoD with the teacher LLM and the standard CoTs distillation method (Std-CoT). We utilize \gtext{\Large\textbf{\ding{51}}} and \rtext{\Large\textbf{\ding{55}}} to denote whether the CoT is correct or incorrect, respectively.

Table \ref{tab:bbh-boolean-expresion} and \ref{tab:bbh-reasoning-color} show that while Std-CoT correctly predicts the final answer on in-domain tasks, it generates incorrect intermediate reasoning steps, indicating that Std-CoT causes student models to capture spurious correlations between questions and answers, without learning to reason with diversity. This is evidenced by the fact that its generated rationales are almost identical to the teacher's CoTs. In contrast, CasCoD produces more distinct CoTs, differing from the teacher's reasoning format, which indicates an enhancement in the student's reasoning generalizability. Tables \ref{tab:sat-math-agieval} and \ref{tab:case-arc} illustrate that on out-of-domain tasks, Std-CoT fails to reason correctly, including logical and factual mistakes, whereas CasCoD can clearly understand the question and provides concise, logically rich reasoning.
\newpage
\label{appendix:case study}
\begin{table*}[!ht]
    \scriptsize
    \centering
    \renewcommand{\arraystretch}{1}
    \begin{tabular}{ll}
    \toprule
    \multicolumn{1}{l}{\textbf{Question:}} &
    \parbox[l]{13cm}{
     Evaluate the result of a random Boolean expression: not False and False or True or False.
    }\\
    \toprule
    \multicolumn{1}{l|}{\textbf{Teacher}}
    &
    \parbox[c]{13cm}{
    Rmember that (i) expressions inside brackets are always evaluated first and that (ii) the order of operations from highest priority to lowest priority is "not", "and", "or", respectively.\\\\
We first simplify this expression "Z" as follows: "Z = not False and False or True or False = A and B or C or D" where "A = not False", "B = False", "C = True", and "D = False". \\\\
Let's evaluate A: A = not False = not (False) = True.\\\\
Plugging in A, B, C, and D, we get: Z = A and B or C or D = True and False or True or False = False or True or False = True.\\\\Therefore, the answer is True.\quad\gtext{\Large\textbf{\ding{51}}}
    } 
    \\\midrule
    \multicolumn{1}{l|}{\textbf{Std-CoT}}
    &
    \parbox[c]{13cm}{
    Rmember that (i) expressions inside brackets are always evaluated first and that (ii) the order of operations from highest priority to lowest priority is "not", "and", "or", respectively.\\\\
    We first simplify this expression "Z" as follows: "Z = not False and False or True or False = A and B or C or D" where "A = not False", "B = False", "C = True", and "D = False".
    \\\\Let's evaluate A: A = not False = not (False) = True.\\\\
Plugging in A, B, C, and D, \rtext{we get: Z = A and B or C or D = True and False or True or False = False or False or False = False.}\\\\Therefore, the answer is True.\quad\rtext{\Large\textbf{\ding{55}}}
    } 
    \\\midrule
    \multicolumn{1}{l|}{\textbf{CasCoD}}
    &
    \parbox[c]{13cm}{
    Removing the brackets first: (not False and False) or (True or False) or False.\\\\
Let's evaluate these expressions one by one:\\\\
- (not False and False): The boolean expression "not False and False" evaluates to True.\\\\
- (True or False): The boolean expression "True or False" evaluates to True.\\\\
- False: The boolean expression False evaluates to False.\\\\
When we evaluate these expressions, we get True.\\\\Therefore, the answer is True.\quad\gtext{\Large\textbf{\ding{51}}}
    } 
    \\
    \bottomrule
    \end{tabular}
    \caption{A boolean-expression case from BIG-Bench Hard.}
    \label{tab:bbh-boolean-expresion}
    \renewcommand{\arraystretch}{1}
\end{table*}

\begin{table*}[!ht]
    \scriptsize
    \centering
    \renewcommand{\arraystretch}{1}
    \begin{tabular}{ll}
    \toprule
    \multicolumn{1}{l}{\textbf{Question:}} &
    \parbox[l]{13cm}{
     Answer extremely simple questions about the colors of objects on a surface. 
Q:On the table, you see several objects arranged in a row: a burgundy paperclip, a purple cup, a mauve pencil, a yellow keychain, and a silver teddy bear. What is the color of the object furthest from the teddy bear?\\
Options: (A) red (B) orange (C) yellow (D) green (E) blue (F) brown (G) magenta (H) fuchsia (I) mauve (J) teal (K) turquoise (L) burgundy (M) silver (N) gold (O) black (P) grey (Q) purple (R) pink
    }\\
    \toprule
    \multicolumn{1}{l|}{\textbf{Teacher}}
    &
    \parbox[c]{13cm}{
    According to this question, the objects are arranged in a row, from left to right, as follows: (1) a burgundy paperclip, (2) a purple cup, (3) a mauve pencil, (4) a yellow keychain, (5) a silver teddy bear.\\\\
The teddy bear is the last item on the list, namely (5). The object furthest from the teddy bear corresponds to (1), which is a burgundy paperclip.\\\\
The color of the paperclip is burgundy. Therefore, the answer is (L)..\quad\gtext{\Large\textbf{\ding{51}}}
    } 
    \\\midrule
    \multicolumn{1}{l|}{\textbf{Std-CoT}}
    &
    \parbox[c]{13cm}{
    According to this question, the objects are arranged in a row, from left to right, as follows: (1) a burgundy paperclip, (2) a purple cup, (3) a mauve pencil, (4) a yellow keychain, (5) a silver teddy bear.
\rtext{The object furthest from the teddy bear is the mauve pencil, which is the third item on the list.}\\\\
The color of the mauve pencil is mauve. Therefore, the answer is (L).\quad\rtext{\Large\textbf{\ding{55}}}
    } 
    \\\midrule
    \multicolumn{1}{l|}{\textbf{CasCoD}}
    &
    \parbox[c]{13cm}{
    Rationale: According to this question, the objects are arranged in a row, from left to right, as follows: (1) a burgundy paperclip, (2) a purple cup, (3) a mauve pencil, (4) a yellow keychain, (5) a silver teddy bear.\\\\
The object furthest from the teddy bear corresponds to (1), which is a burgundy paperclip.\\\\
The color of the paperclip is burgundy.\\\\
Therefore, the answer is (L).\quad\gtext{\Large\textbf{\ding{51}}}
    } 
    \\
    \bottomrule
    \end{tabular}
    \caption{A reasoning about colored objects case from BIG-Bench Hard.}
    \label{tab:bbh-reasoning-color}
    \renewcommand{\arraystretch}{1}
\end{table*}

\begin{table*}[!ht]
    \scriptsize
    \centering
    \renewcommand{\arraystretch}{1}
    \begin{tabular}{ll}
    \toprule
    \multicolumn{1}{l}{\textbf{Question:}} &
    \parbox[l]{13cm}{
     $h=3 a+28.6$ A pediatrician uses the model above to estimate the height $h$ of a boy, in inches, in terms of the boy's age $a$, in years, between the ages of 2 and 5. Based on the model, what is the estimated increase, in inches, of a boy's height each year?\\Options:
(A)3 (B)5.7 (C)9.5 (D)14.3
    }\\
    \toprule
    \multicolumn{1}{l|}{\textbf{Teacher}}
    &
    \parbox[c]{13cm}{
    Step 1: We are given the model $h=3a+28.6$ which represents the estimated height $h$ of a boy in terms of his age $a$.\\\\
Step 2: We need to find the estimated increase in height each year. This means we need to find the change in height for a change in age of 1 year.\\\\
Step 3: To find the change in height, we can subtract the height at age $a+1$ from the height at age $a$. So, the change in height is $(3(a+1)+28.6) - (3a+28.6)$.\\\\
Step 4: Simplifying the expression, we get $3a+3+28.6 - 3a - 28.6$.\\\\
\rtext{Step 5: The terms $3a$ and $-3a$ cancel out, leaving us with $3+3$.}\\\\
Step 6: The change in height is $6$ inches.\\\\
Therefore, the answer is (A) 3.\quad\rtext{\Large\textbf{\ding{55}}}
    } 
    \\\midrule
    \multicolumn{1}{l|}{\textbf{Std-CoT}}
    &
    \parbox[c]{13cm}{
    This question focuses on the increase in height each year. \rtext{According to the model, the height of a boy between the ages of 2 and 5 is approximately 36 inches.} Therefore, the estimated increase in height each year is 3 inches. Therefore, the answer is (A).\quad\rtext{\Large\textbf{\ding{55}}}
    } 
    \\\midrule
    \multicolumn{1}{l|}{\textbf{CasCoD}}
    &
    \parbox[c]{13cm}{
    Rationale: This question asks about the estimated increase, in inches, of a boy's height each year. According to the model, the height of a boy increases by 3 inches each year between the ages of 2 and 5.\\\\
Therefore, the answer is (A) 3 inches.\quad\gtext{\Large\textbf{\ding{51}}}
    } 
    \\
    \bottomrule
    \end{tabular}
    \caption{A sat-math case from AGIEval.}
    \label{tab:sat-math-agieval}
    \renewcommand{\arraystretch}{1}
\end{table*}

\begin{table*}[!ht]
    \scriptsize
    \centering
    \renewcommand{\arraystretch}{1}
    \begin{tabular}{ll}
    \toprule
    \multicolumn{1}{l}{\textbf{Question:}} &
    \parbox[l]{13cm}{
     Which gas is the most abundant in Earth's atmosphere?\\Options:
(A) oxygen (B) nitrogen (C) water vapor (D) carbon dioxide
    }\\
    \toprule
    \multicolumn{1}{l|}{\textbf{Teacher}}
    &
    \parbox[c]{13cm}{
    First, we know that Earth's atmosphere is primarily composed of gases. \\\\
Second, we need to determine which gas is the most abundant. \\\\
Third, we can eliminate options (A) oxygen and (C) water vapor because they are present in smaller quantities compared to the other gases. \\\\
Finally, we are left with options (B) nitrogen and (D) carbon dioxide. \\\\
Considering that nitrogen makes up about 78\% of Earth's atmosphere, while carbon dioxide only makes up about 0.04\%, it is clear that nitrogen is the most abundant gas in Earth's atmosphere. \\\\
Therefore, the answer is (B) nitrogen.\quad\gtext{\Large\textbf{\ding{51}}}
    } 
    \\\midrule
    \multicolumn{1}{l|}{\textbf{Std-CoT}}
    &
    \parbox[c]{13cm}{
   The percentage of each gas in the atmosphere is given by the following table:\\\\
Gas Percentage\\\\
oxygen 20.95\\\\
nitrogen 78.09\\\\
water vapor 0.4\\\\
carbon dioxide 0.04\\\\
\rtext{According to this table, oxygen is the most abundant gas in Earth's atmosphere.} Therefore, the answer is (A).\quad\rtext{\Large\textbf{\ding{55}}}
    } 
    \\\midrule
    \multicolumn{1}{l|}{\textbf{CasCoD}}
    &
    \parbox[c]{13cm}{
    Rationale: According to the World Bank, the composition of Earth's atmosphere is approximately 78\% nitrogen, 21\% oxygen, 0.93\% argon, and small amounts of other gases. Nitrogen is the most abundant gas in Earth's atmosphere.\\\\
Therefore, the answer is (B).\quad\gtext{\Large\textbf{\ding{51}}}
    } 
    \\
    \bottomrule
    \end{tabular}
    \caption{A case from AI2 Reasoning Challenge.}
    \label{tab:case-arc}
    \renewcommand{\arraystretch}{1}
\end{table*}
\end{document}